\documentclass[graybox]{svmult}

\usepackage{mathptmx}       
\usepackage{helvet}         
\usepackage{courier}        
\usepackage{type1cm}        
\usepackage{makeidx}         
\usepackage{graphicx}        
\usepackage{multicol}        
\usepackage[bottom]{footmisc}
\usepackage{url}

\usepackage{ifthen}
\usepackage{makeglos}

\graphicspath{{./}{./graphics/}{./graphics/eps/}{./graphics/pdf/}}

\usepackage{subfigure}
\usepackage{soul}

\newcommand{\buildbook}{false}

\makeindex             

\makeglossary

\begin{document}

\title*{Introduction to Presentation Attacks in Signature Biometrics and Recent Advances}


\author{Carlos Gonzalez-Garcia, Ruben Tolosana, Ruben Vera-Rodriguez, \\ Julian Fierrez and Javier Ortega-Garcia}
\institute{Carlos Gonzalez-Garcia \at Universidad Autonoma de Madrid, Spain, \email{carlos.gonzalezgarcia@estudiante.uam.es}
\and Ruben Tolosana \at Universidad Autonoma de Madrid, Spain, \email{ruben.tolosana@uam.es}
\and Ruben Vera-Rodriguez \at Universidad Autonoma de Madrid, Spain, \email{ruben.vera@uam.es}
\and Julian Fierrez \at Universidad Autonoma de Madrid, Spain, \email{julian.fierrez@uam.es}
\and Javier Ortega-Garcia \at Universidad Autonoma de Madrid, Spain, \email{javier.ortega@uam.es}}

%

%
\maketitle

%


\abstract*{
Applications based on biometric authentication have
received a lot of interest in the last years due to the breathtaking
results obtained using personal traits such as face or fingerprint. However, it is important not to forget that these biometric systems have to withstand different types of possible attacks. This chapter carries out an analysis of different Presentation Attack (PA) scenarios for on-line handwritten signature verification. The main contributions of this chapter are: \textit{i)} an updated overview of representative methods for Presentation Attack Detection (PAD) in signature biometrics; \textit{ii)} a description of the different levels of PAs existing in on-line signature verification regarding the amount of information available to the impostor, as well as the training, effort, and ability to perform the forgeries; and \textit{iii)} an evaluation of the system performance in signature biometrics under different scenarios considering recent publicly available signature databases, DeepSignDB\footnote{\url{https://github.com/BiDAlab/DeepSignDB}} and SVC2021\_EvalDB\footnote{\url{https://github.com/BiDAlab/SVC2021\_EvalDB}},\footnote{\url{https://codalab.lisn.upsaclay.fr/competitions/9189}}. This work is in line with recent efforts in the Common Criteria standardization community towards security evaluation of biometric systems.}

\abstract{
Applications based on biometric authentication have
received a lot of interest in the last years due to the breathtaking
results obtained using personal traits such as face or fingerprint. However, it is important not to forget that these biometric systems have to withstand different types of possible attacks. This chapter carries out an analysis of different Presentation Attack (PA) scenarios for on-line handwritten signature verification. The main contributions of this chapter are: \textit{i)} an updated overview of representative methods for Presentation Attack Detection (PAD) in signature biometrics; \textit{ii)} a description of the different levels of PAs existing in on-line signature verification regarding the amount of information available to the impostor, as well as the training, effort, and ability to perform the forgeries; and \textit{iii)} an evaluation of the system performance in signature biometrics under different scenarios considering recent publicly available signature databases, DeepSignDB\footnote{\url{https://github.com/BiDAlab/DeepSignDB}} and SVC2021\_EvalDB\footnote{\url{https://github.com/BiDAlab/SVC2021\_EvalDB}}. This work is in line with recent efforts in the Common Criteria standardization community towards security evaluation of biometric systems.}

\section{Introduction}

Signature verification systems have become very popular in many applications such as banking, e-health, e-education, and security in recent times \cite{faundez2020handwriting}. This evolution has been motivated due to two main factors: \textit{i}) the technological evolution and the improvement of sensors quality, which has made general purpose devices (smartphones \cite{eBioSign_journal} and tablets \cite{Alonso-Fernandez2005_TabletPC}) more accessible to the general population, and therefore, the social acceptance has increased; and \textit{ii}) the evolution of biometric recognition technologies, especially through the use of deep learning techniques \cite{diaz2019perspective, 2021_TBIOM_DeepSign_Tolosana}. However, it is important to highlight that this biometric systems have to endure different types of possible attacks \cite{Galbally2007_Vulnerabilities, 2021_Book_DigitalFaceManipulation}, some of them highly complex \cite{hadid15SPMspoofing}.

In this chapter we focus on the study of different Presentation Attack (PA) scenarios for on-line handwritten signature biometric verification systems due to the significant amount of attention received in the last years thanks to the development of new scenarios (e.g. device interoperability \cite{2015_IEEEAccess_InterSign_Tolosana} and mobile scenarios \cite{mobile_scenario, local_features}) and writing tools (e.g. finger \cite{eBioSign_journal}). These new scenarios have grown hand in hand with the rapid expansion of mobile devices, such as smartphones, which allow the implementation of biometric-based verification systems far from the traditional office-like ones \cite{MobileTouchDB}. 

In general, two different types of impostors can be found in the context of signature verification: 1) \textit{random (zero-effort} or \textit{accidental)} impostors, the case in which no information about the signature of the user being attacked is known and impostors present their own genuine signature claiming to be another user of the system, and 2) \textit{skilled} impostors, the case in which attackers have some level of information about the signature of the user to attack (e.g. global shape of the signature or signature dynamics) and try to forge the signature claiming to be that user in the system.

In \cite{Galbally_IJCB_2017}, Galbally \textit{et al.} discussed different approaches to report accuracy results in handwritten signature verification. They considered skilled impostors as a particular case of biometric PAs which is performed against a behavioural biometric characteristic (also referred as \textit{mimicry}). There are important differences between PAs and mimicry: while traditional PAs involve the use of some physical artefacts such as fake masks and gummy fingers (and therefore, they can be detected in some cases at the sensor level), in the case of mimicry the interaction with the sensor is exactly the same followed in a genuine access attempt. In \cite{Galbally_IJCB_2017} a different nomenclature of impostor scenarios is proposed following the literature standard in the field of biometric Presentation Attack Detection (PAD) \glossary{PAD: Presentation Attack Detection}: the classical random impostor scenario is referred to as Bona Fide (BF) scenario, while the skilled impostor scenario is referred to as PA \glossary{PA: Presentation Attack} scenario. This nomenclature has also been used in this chapter.   

If during the development of a biometric verification system those PAs are expected, it is possible to include specific modules for PAD, which in the signature verification literature are commonly referred to as forgery detection modules. A comprehensive study of these PAD methods is out of the scope of the chapter, but in Sec. \ref{sec:introduction_PA} we provide a brief overview of some selected representative works in that area.

A different approach to improve the security of a signature verification system against attacks different from including a PAD module is template protection \cite{2010_SMC_A_Campisi, delgado2019biometric, Freire2008_ICASSP, 2017_Access_HEmultiDTW_Marta, GOMEZBARRERO2017149, NANNI20103676, ponce2020fuzzy}. Traditional on-line signature verification systems work with very sensitive biometric data such as the \textit{X} and \textit{Y} spatial coordinates and store that information without any additional protection. This makes very easy for attackers to steal this information. If an attacker has the information of spatial coordinates along the time axis it would be very easy for him/her to generate very high quality forgeries. Template protection techniques involve feature transformation and the use of biometric cryptosystems. In \cite{2015_ICCST_SignatureRobTemp_RubenT}, an extreme approach for signature template generation was proposed not considering information related to \textit{X}, \textit{Y} coordinates and their derivatives on the biometric system, providing therefore a much more robust system against attacks, as this critical information would not be stored anywhere. Moreover, the results achieved had error rates in the same range as more traditional systems which store very sensitive information. An interesting review and classification of different biometric template protection techniques for on-line handwritten signature application is conducted in \cite{malallah2013review}.

The main contributions of this chapter are: \textit{i)} a brief overview of representative methods for PAD in signature biometrics; \textit{ii)} a description of the different levels of PAs existing in on-line signature verification regarding the amount of information available to the impostor, as well as the training, effort and ability to perform the forgeries; and \textit{iii)} an evaluation of the system performance in signature biometrics under different scenarios following the recent SVC-onGoing competition\footnote{\url{https://codalab.lisn.upsaclay.fr/competitions/9189}}  \cite{tolosana2021icdar}.

The remainder of the chapter is organized as follows. The introduction is completed with a short overview of PAD in signature biometrics (Sec. \ref{sec:introduction_PA}). After that, the main technical content of the chapter begins in Sec. \ref{sec:impostor_scenarios}, with a review of the most relevant features of all different impostor scenarios, pointing out which type of impostors are included in many different well-known public signature databases. Sec. \ref{sec:databases} describes the on-line signature databases considered in the experimental work. Sec. \ref{sec:experimental_work} describes the experimental protocol and the results achieved. Finally, Sec. \ref{sec:conclusions} draws the final conclusions and points out some lines for future work.

\section{Review of PAD in Signature Biometrics}\label{sec:introduction_PA}

Presentation Attack Detection (PAD) in signature biometrics is a field that has been extensively studied since the late 70s to the present \cite{1977_Rosenfeld}. In this section we describe some state-of-the-art forgery detection methods.

Some of the studies that can be found in the literature are based on the \textit{Kinematic Theory} of rapid human movements and its associated Sigma LogNormal model. In \cite{2015_ICB_skilledSignSigmaLog_Marta}, the authors proposed a new scheme in which a module focused on the detection of skilled forgeries (i.e. PA impostors) was based on four parameters of the Sigma LogNormal writing generation model \cite{Plamondon_sigmaLogNormal} and a linear classifier. That new binary classification module was supposed to work sequentially before a standard signature recognition system \cite{2018_INFFUS_MCSreview1_Fierrez}. Good results were achieved using that approach for both skilled (i.e. PA) and random (i.e. BF) scenarios. In \cite{Reillo_PAs}, Reillo \textit{et al.} proposed PAD methods based on the use of some global features such as the total number of strokes and the signing time of the signatures. They acquired a new database based on 11 levels of PAs regarding the level of knowledge and the tools available to the forger. The results achieved in that work using the proposed PAD methods reduced the Equal Error Rate (EER) \glossary{EER: Equal Error Rate} from a percentage close to 20.0\% to below 3.0\%.

In \cite{Lovell_detection}, authors proposed an off-line signature verification and forgery detection system based on fuzzy modelling. The verification of genuine signatures and detection of forgeries was achieved via angle features extracted using a grid method. The derived features were fuzzified by an exponential membership function, which was modified to include two structural parameters regarding variations of the handwriting styles and other factors affecting the scripting of a signature. Experiments showed the capability of the system in detecting even the slightest changes in signatures. 

Brault and Plamondon presented in \cite{Plamondon_1993_modellingForgeries} an original attempt to estimate, quantitatively and a priori from the coordinates sampled during its execution, the difficulty that could be experienced by a typical imitator in reproducing both visually and dynamically that signature. To achieve this goal, they first derived a functional model of what a typical imitator must do to copy dynamically any signature. A specific difficulty coefficient was then numerically estimated
for a given signature. Experimentation geared specifically to signature imitation demonstrated the effectiveness of the model. The ranking of the tested signatures given by the difficulty coefficient was compared to three different sources: the opinions of the imitators themselves, the ones of an expert document examiner, and the ranking given by a specific pattern recognition algorithm. They provided an example of application as well. This work was one of the first attempts of PAD for on-line handwritten signature verification using a special pen attached to a digitizer (Summagraphic Inc. model MM1201). The sampling frequency was 110 Hz, and the spatial resolution was 0.025 inch. 

Finally, it is important to highlight that new approaches based on deep learning architectures are commonly used in the literature \cite{2021_TBIOM_DeepSign_Tolosana, tolosana2021icdar}. Several studies use Convolutional Neural Network (CNN) \glossary{CNN: Convolutional Neural Network} architectures in order to predict whether a signature is genuine or a forgery presented by a PA impostor \cite{wu2019deep, vorugunti2019osvnet, lai2021synsig2vec2}. Other state-of-the-art architectures are based on Recurrent Neural Networks (RNNs) \glossary{RNN: Recurrent Neural Network} such as the ones presented in \cite{2018_IEEEAccess_RNN_Tolosana}. Also, some recent works focus on analyzing the system performance against both BF and PA scenarios depending of the signature complexity \cite{2021_ICPRw_Exploit_Signature_Complexity, 2019_ICDAR_DeepSignCX_Vera}.

\section{Presentation Attacks in Signature Biometrics}\label{sec:impostor_scenarios}

The purpose of this section is to clarify the different levels of skilled forgeries (i.e. PA impostors) that can be found in the signature biometrics literature regarding the amount of information provided to the attacker, as well as the training, effort and ability to perform the forgeries. In addition, the case of random forgeries (i.e. zero-effort impostors) is also considered although it belongs to the BF scenario and not to the PA scenario in order to review the whole range of possible attacks in on-line signature verification.

Previous studies have applied the concept of Biometric Menagerie in order to categorize each type of user of the biometric system as an animal. This concept was initially formalized by Doddington \textit{et al.} in \cite{Doddington}, classifying speakers regarding the ease or difficulty with which the speaker can be recognized (i.e. sheep and goats, respectively), how easily they can be forged (i.e. lambs) and finally, how adept/effective they are at forging/imitating the voice of others (i.e. wolves). Yager and Dunstone extended the Biometric Menagerie in \cite{Yager_biometricMenagerie} by adding four more categories of users (i.e. worms, chameleons, phantoms, and doves). Their proposed approach was investigated using a broad range of biometric modalities, including 2D and 3D faces, fingerprints, iris, speech, and keystroke dynamics. In \cite{Sonia_PlosOne}, Houmani and Garcia-Salicetti applied the concept of Biometric Menagerie for the different types of users found in the on-line signature verification task proposing the combination of their personal and relative entropy measures as a way to quantify how difficult it is a signature to be forged. Their proposed approach achieved promising classification results on the MCYT database \cite{Ortega_Garcia2003_MCYT}, where the attacker had access to a visual static image of the signature to forge.

In \cite{Ballard_quality_2007}, some experiments were carried out to reach the following conclusions: 1) some users are significantly better forgers than others; 2) forgers can be trained in a relatively straight-forward way to become a greater threat; 3) certain users are easy targets for forgers; and 4) most humans are relatively poor judges of handwriting authenticity, and hence, their unaided instincts cannot be trusted. Additionally, in that work authors proposed a new metric for impostor classification more realistic to the definition of security, i.e., \textit{naive}, \textit{trained}, and \textit{generative}. They considered naive impostors as random impostors (i.e. zero-effort impostors) in which no information about the user to forge is available whereas they referred to trained and generative impostors to skilled forgeries (i.e. PA impostors) when only the image or the dynamics of the signature to forge is available, respectively.

In \cite{test_tool_2003}, the authors proposed a software tool implemented on two different computer platforms in order to achieve forgeries with different quality levels (i.e. PA impostors). Three different levels of PAs were considered: 1) \textit{blind forgeries}, the case in which the attacker writes on a blank surface having access just to textual knowledge (i.e. precise spelling of the user's name to forge); 2) \textit{low-force forgeries}, where the attacker gets a blueprint of the signature projected on the writing surface (dynamic information is not provided), which they may trace; and 3) \textit{brute-force forgeries}, in which an animated pointer is projected onto the writing pad showing the whole realization of the signature to forge. The attacker may observe the sequence and follow the pointer. The authors carried out an experiment based on the use of 82 forgery samples performed by four different users in order to detect how the False Acceptance Rate (FAR) is affected regarding the level of PA. They considered a signature verification system based on the average quadratic deviation horizontal and vertical writing signals. Results obtained for four different threshold values confirmed the requirement of strong protection of biometric reference data as it was proposed in \cite{2015_ICCST_SignatureRobTemp_RubenT}.


\subsection{Types of Presentation Attacks}

Alonso-Fernandez \textit{et al.} carried out an exhaustive analysis of the different types of forgeries found in handwritten signature verification systems \cite{Fernando_robustness_2009}. In that work, authors considered random impostors and 4 different levels of PA impostors, classified regarding the amount of information provided and the tools used by the attacker in order to forge the signature:

\begin{itemize}
\item \textbf{Random or zero-effort forgeries}, in which no information of the user to forge is available and the attacker uses its own genuine signature (accidentally or not) claiming to be another user of the system. 

\item \textbf{Blind forgeries}, in which the impostor has access to a descriptive or textual knowledge of signatures to forge (e.g. the name of the subject to forge).

\item \textbf{Static forgeries} (low-force in \cite{test_tool_2003}), where the attacker has available a static image of the global shape of the signature to forge. In this case, there are two ways to generate the forgeries.  In the first one, the attacker can train to imitate the signature with or without time restrictions and blueprint, and then forge it without the use of the blueprint, which leads to \textbf{static trained forgeries}. In the second one, the attacker uses a blueprint to first copy the genuine signature of the user to forge and then put it on the screen of the device while forging, leading to \textbf{static blueprint forgeries}, more difficult to detect as they have quite the same appearance as the original ones.

\item \textbf{Dynamic forgeries} (brute-force in \cite{test_tool_2003}), where the impostor has access to both the global image and also the whole realization process (i.e. dynamics) of the signature to forge. The dynamics can be obtained in the presence of the original writer or through the use of a video-recording. In a similar way as the previous category, we can distinguish first \textbf{dynamic trained forgeries} in which the attacker can use specific tools to analyze and train to forge the genuine signature, and second, \textbf{dynamic blueprint forgeries} which are generated by projecting on the acquisition area a real-time pointer that the forger only needs to follow. 

\item \textbf{Regained forgeries}, the case where the impostor has only available the static image of the signature to forge and makes use of a dedicated software to recover the signature dynamics \cite{aythami_sintesis_2017}, which are later analyzed and used to create dynamic forgeries.
\end{itemize}

\begin{figure}[tb]
\centering
\subfigure[Genuine Signature]{
\includegraphics[width=0.45\linewidth]{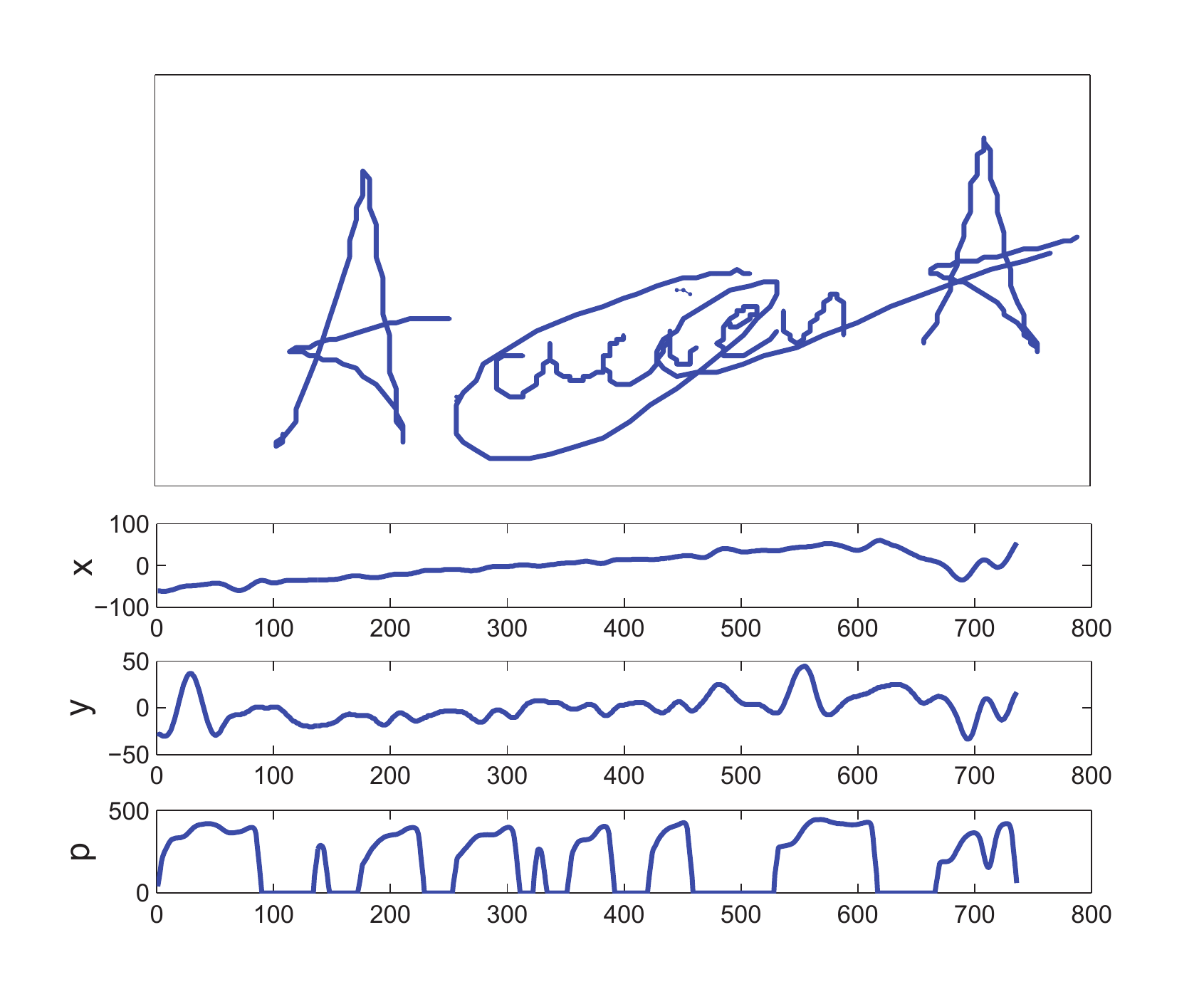}}
\subfigure[Random Forgery]{
\includegraphics[width=0.45\linewidth]{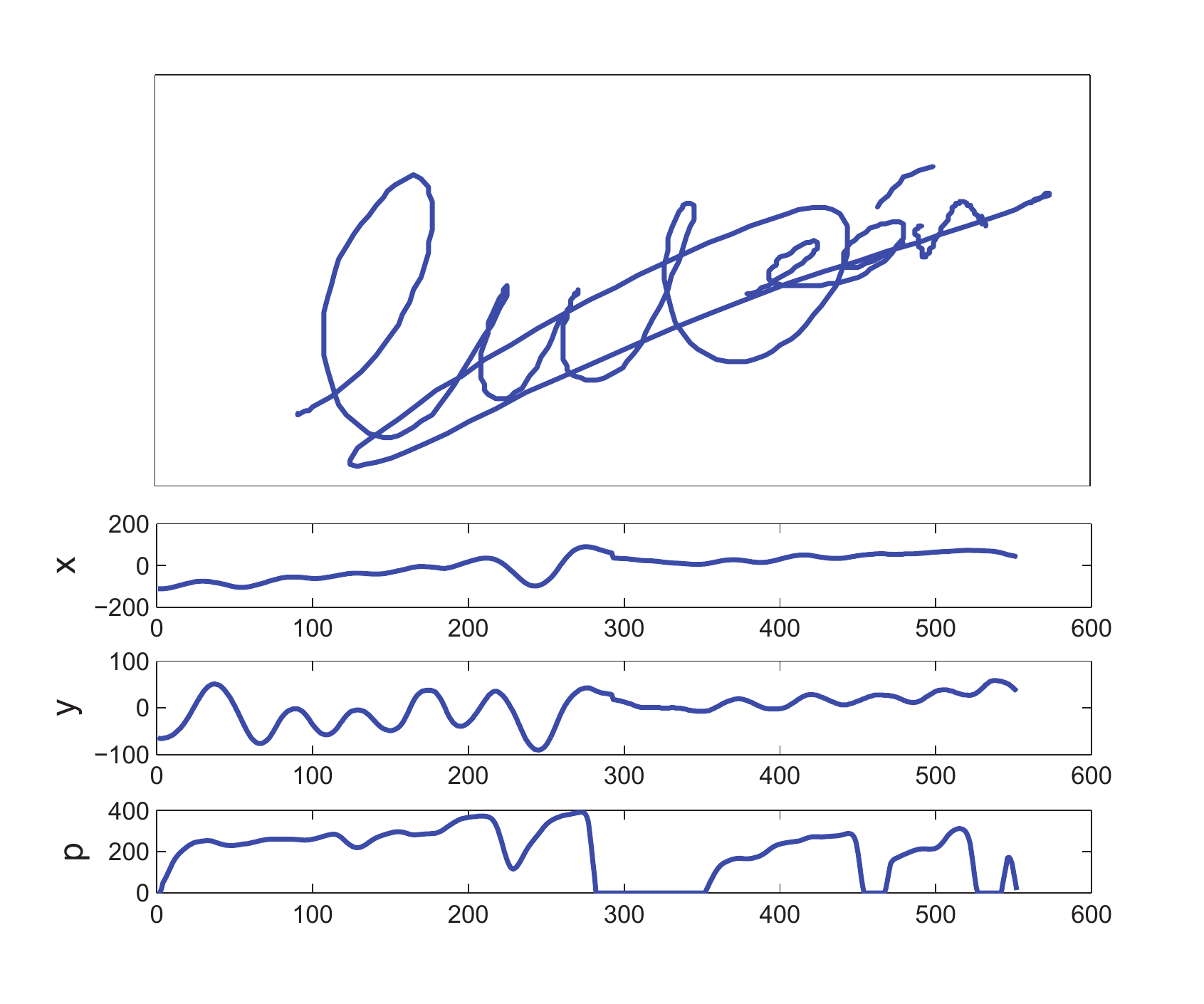}}
\subfigure[Static Blueprint Forgery]{
\includegraphics[width=0.46\linewidth]{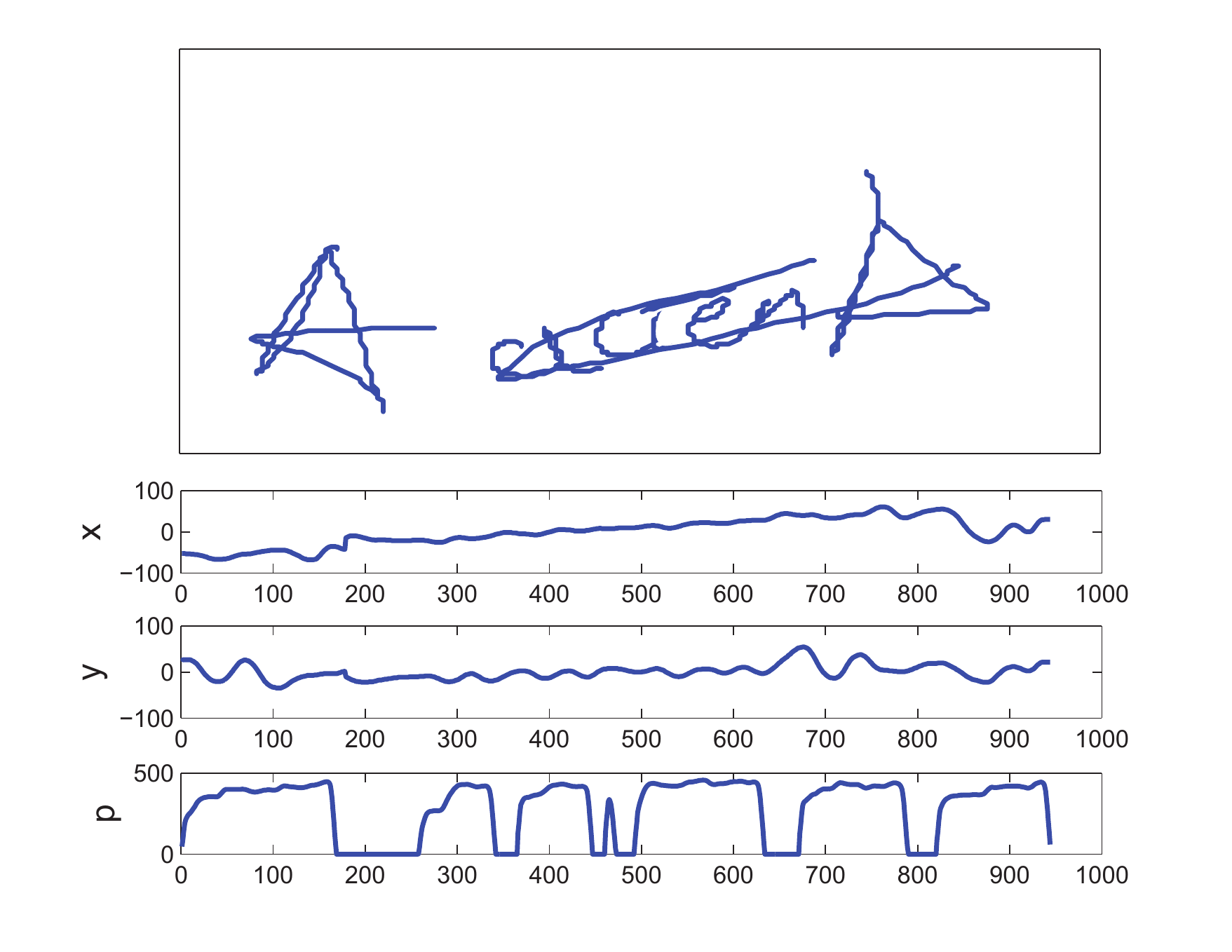}}
\hspace{0.2cm}
\subfigure[Dynamic Trained Forgery]{
\includegraphics[width=0.45\linewidth]{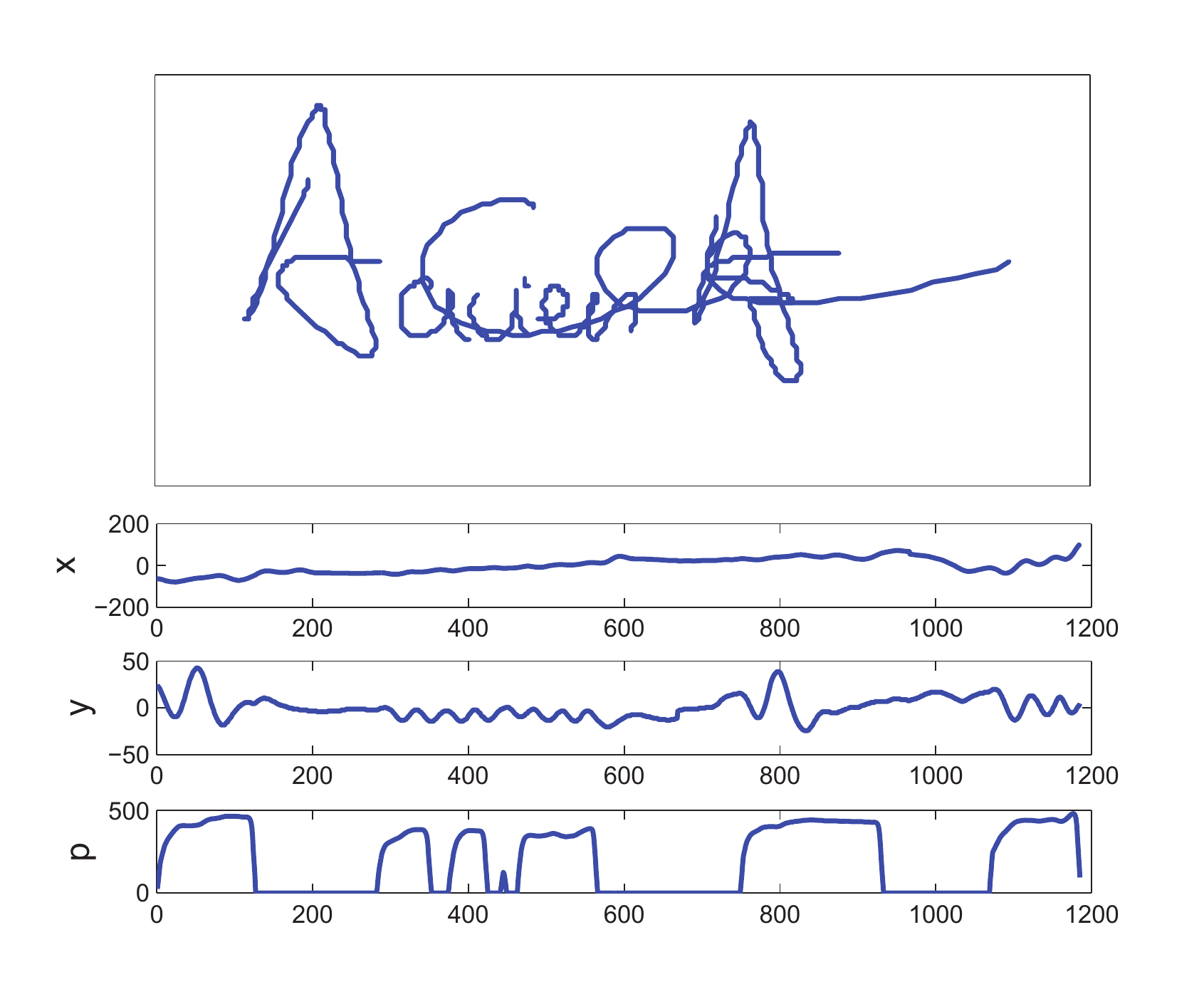}}
\caption{Examples of one genuine signature and three different types of forgeries performed for the same user.} \label{types_forgeries}
\end{figure}

As expected, dynamic forgeries are the forgeries with better quality (in some cases, very similar to the genuine signatures that are forged), followed by static forgeries. Random and blind forgeries are usually very different from the signature forged. Fig. \ref{types_forgeries} shows examples of a genuine signature and three different types of forgeries (i.e. random, static blueprint and dynamic trained) performed for the same user. The image shows both the static and dynamic information with the \textit{X} and \textit{Y} coordinates and pressure.

\begin{figure*}[t]
  \centering
    \includegraphics[width=\linewidth]{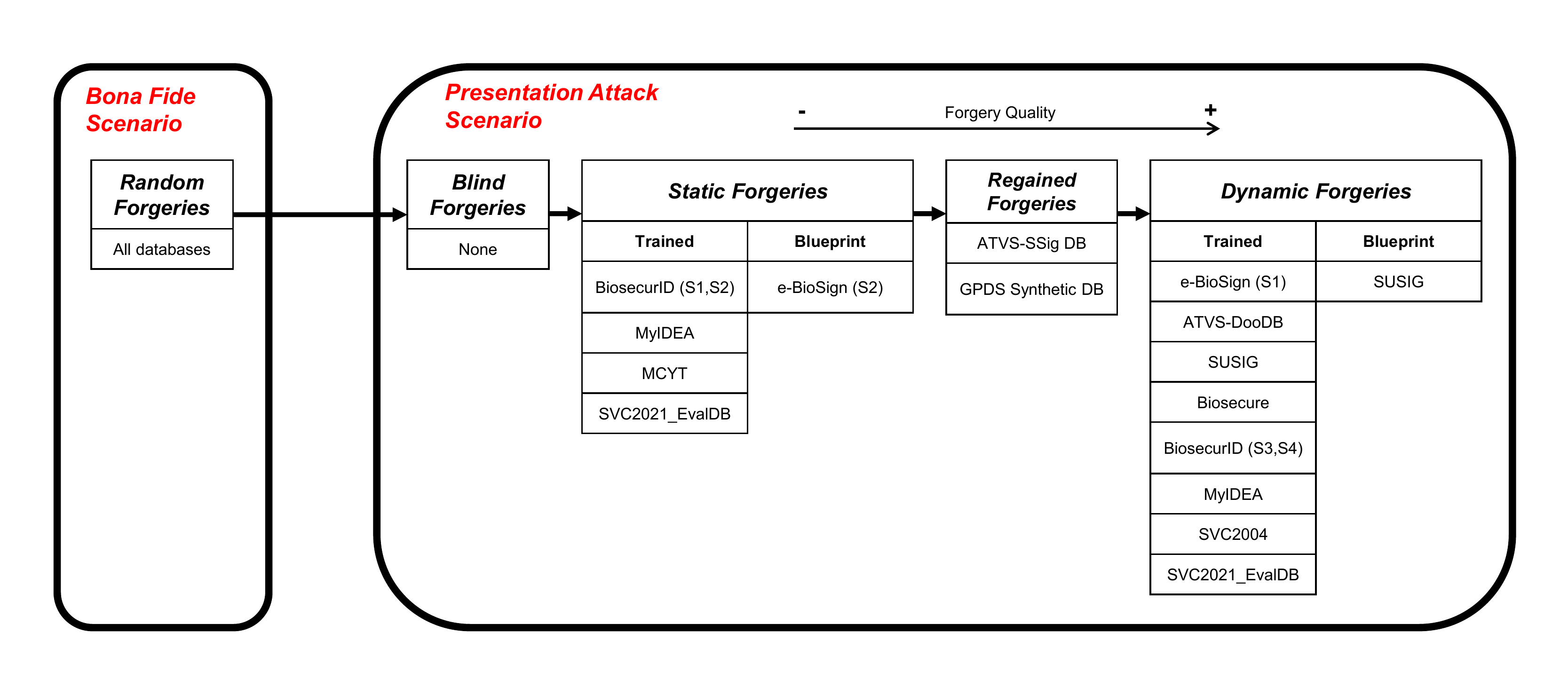}
  \caption{Diagram of different types of forgeries for both BF and PA scenarios regarding the amount of information provided to the attacker, as well as the training, effort, and ability to perform them. The most commonly used on-line signature databases are included to each PA group.}
  \label{fig:skilled_complexity}
\end{figure*}

Besides the forgery classification carried out in \cite{Fernando_robustness_2009}, Alonso-Fernandez \textit{et al.} studied the impact of an incremental level of quality forgeries against handwritten signature verification systems. The authors considered off-line and on-line systems using the BiosecurID database \cite{Fierrez2009_PAA}. For the off-line verification system, they considered a system based on global image analysis and a minimum distance classifier \cite{Fierrez-Aguilar2004_BIOAW} whereas a system based on Hidden Markov Models (HMM) \cite{2015_WIFS_SignatureHMMUpdate} was considered for the on-line system. The experiments carried out proved that the performance of the off-line approach is only degraded when the highest quality level of forgeries is used. The on-line system shows a progressive degradation of its performance when the quality level of the forgeries is increased. This lead the authors to the conclussion that the dynamic information of signatures is the one more affected when the quality of the forgeries increases.

Finally, Fig. \ref{fig:skilled_complexity} summarizes all different types of forgeries for both BF and PA scenarios regarding the amount of information provided to the impostor, as well as the training, effort, and ability to perform them. In addition, the most commonly used on-line signature databases are included to each PA group in order to provide an easy representation. To the authors' best knowledge, there are no publicly available on-line signature databases fot the case of blind forgeries.

\subsection{Synthetic Forgeries}

On-line signature synthesis has become a very interesting research line due to, among other reasons,  the lack of forgery signatures in real scenarios, which makes the development of robust signature verification systems difficult \cite{diaz2019perspective}.

One of the most popular approaches in the literature for realistic handwriting generation was presented in \cite{graves2013generating}. In that study, the author presented a Long Short-Term Memory (LSTM) Recurrent Neural Network (RNN) architecture to generate complex sequences. The proposed architecture was tested on handwriting, achieving very good visual results. 
Currently, the Sigma LogNormal model is one of the most popular on-line signature synthesis approaches \cite{ferrer2018idelog, 2020_BookLogNormal_ComplexSignTouch_Vera}, and has been applied to on-line signature verification systems, generating synthetic samples from a genuine signature, increasing the amount of information and improving the performance of the systems \cite{Moises_Tran_Cybernetics, galbally09ICDARenrollment, lai2020synsig2vec}.

Other important contributions in this area are the following ones. In \cite{aythami_sintesis_2017}, Ferrer \textit{et al.} proposed a system for the synthetic generation of dynamic information for both static and dynamic handwritten signatures based on the motor equivalence theory, which divides the action of human handwiting into an effector dependent cognitive level and an effector independent motor level, achieving good results. In \cite{tolosana2021deepwritesyn}, Tolosana \textit{et al.} proposed DeepWriteSYN, a novel on-line handwriting signature synthesis approach based on deep short-term representations. The DeepWriteSYN architecture is composed by two modules: a first module which divides the signature in short-time strokes and a second module based on a sequence-to-sequence Variational Autoencoder (VAE) in charge of the synthesis of those short-time strokes. DeepWriteSYN is able to generate realistic handwriting variations of a given handwritten structure corresponding to the natural variation within a given population or a given subject. For more information, an exhaustive study of the evolution of synthetic handwriting is conducted in \cite{carmona2017temporal}.

\section{On-Line Signature Databases}\label{sec:databases}
The following two public databases are considered in the experiments reported here. Both of them are currently used in the popular SVC-onGoing on-line signature verification competition\footnote{\url{https://codalab.lisn.upsaclay.fr/competitions/9189}}.

\subsection{DeepSignDB}
The DeepSignDB\footnote{\url{https://github.com/BiDAlab/DeepSignDB}} database \cite{DeepSignDB} is composed by a total of 1,526 subjects from four different well known state-of-the-art databases: MCYT (330 subjects) \cite{Ortega_Garcia2003_MCYT}, BiosecurID (400 subjects) \cite{Fierrez2009_PAA}, Biosecure DS2 (650 subjects) \cite{Houmani2012993}, e-BioSign (65 subjects) \cite{eBioSign_journal}, and a novel on-line signature database composed by 81 subjects. DeepSignDB comprises more than 70K signatures acquired using both stylus and finger writing inputs in both office and mobile scenarios. A total of 8 different devices were considered during the acquisition process (i.e., 5 Wacom devices and 3 Samsung general purpose devices). In addition, different types of impostors and number of acquisition sessions are considered along the database.

The available information when using the pen stylus as writing input is \textit{X} and \textit{Y} spatial coordinates and pressure. In addition, pen-up trajectories are also available. For the case of using the finger as writing input, the only available information is \textit{X} and \textit{Y} spatial coordinates.

\subsection{SVC2021\_EvalDB}
The SVC2021\_EvalDB \footnote{\url{https://github.com/BiDAlab/SVC2021\_EvalDB}} is a novel database specifically acquired for the ICDAR 2021 Signature Verification Competition \cite{tolosana2021icdar} and then used as well for the SVC-onGoing Competition \cite{tolosana2021icdar}. In this database, two scenarios are considered: office and mobile scenarios.

\begin{itemize}
\item \textbf{Office scenario:} on-line signatures from 75 subjects were collected using a Wacom STU-530 device with the stylus as writing input. It is important to highlight that all the acquisition took place in an office scenario under the supervision of a person with experience in the on-line signature verification field. The subjects considered in the acquisition of SVC2021\_EvalDB database are different compared to the ones considered in the previous DeepSign database. All the signatures were collected in two different sessions separated by at least 1 week. For each genuine subject, a total of 8 genuine signatures (4 genuine signatures per session) and 16 skilled forgeries (8 static forgeries and 8 dynamic forgeries, performed by 4 different subjects in two different sessions) were collected. Regarding the skilled forgeries, static forgeries were collected in the first acquisition session and dynamic forgeries were considered in the second one. The following information is available for every signature: \textit{X} and \textit{Y} spatial coordinates, pressure, pen-up trajectories and timestamp.

\item \textbf{Mobile scenario:} on-line signatures from a total of 119 subjetcts were acquired using the same acquisition framework considered in MobileTouchDB database \cite{MobileTouchDB}: an Android App was developed in order to work with unsupervised mobile scenarios. All users could download the application and use it on their own smartphones without any kind of supervision, simulating a real scenario (e.g., standing, sitting, walking, in public transport, etc.). As a result, a total of 94 different smartphone models from 16 different brands are available in the database.
Regarding the acquisition protocol, between four and six separated sessions were acquired for every user with a time gap between first and last session of at least 3 weeks. The number and type of the signatures for every user is the same as on the office scenario. Timestamp and spatial coordinates \textit{X} and \textit{Y} are available for every signature.
\end{itemize}

\section{Experimental Work}\label{sec:experimental_work}

\subsection{On-line Signature Verification System}\label{sec:signature_verification}

We consider for the experimental analysis the state-of-the-art signature verification system presented in \cite{MobileTouchDB, 2021_TBIOM_DeepSign_Tolosana} based on Time-Alignment Recurrent Neural Network (TA-RNN) \glossary{TA-RNN: Time Alignment Recurrent Neural Network}.

For the input of the system, the network is fed with 23 time functions extracted from the signature \cite{martinez14mobileSignRobustPerf}. Information related to the azimuth and altitude of the pen angular orientation is not considered in this case. The TA-RNN architecture is based on two consecutive stages: \textit{i)} time sequence alignment through DTW (\textit{Dynamic Time Warping}) \glossary{DTW: Dynamic Time Warping}, and \textit{ii)} feature extraction and matching using a RNN. The RNN system comprises three layers. The first layer is composed of two Bidirectional Gated Recurrent Unit (BGRU) hidden layers with 46 memory blocks each, sharing the weights between them. The outputs of the first two parallel BGRU hidden layers are concatenated and serve as input to the second layer, which corresponds to a BGRU hidden layer with 23 memory blocks. Finally, a feed-forward neural network layer with a sigmoid activation is considered, providing an output score for each pair of signatures. This learning model was presented in \cite{2021_TBIOM_DeepSign_Tolosana} and was retrained for the SVC-onGoing competition \cite{tolosana2021icdar} adapted to the stylus scenario by using only the stylus-written signatures of the development set of DeepSignDB (1,084 users). The best model has been then selected using a partition of the development set of DeepSignDB, leaving out of the training the DeepSignDB evaluation set (442 users).

\subsection{Experimental Protocol}\label{sec:experimentalProtocol}

The experimental protocol has been designed to allow the study of both random forgeries (i.e. BF) and skilled forgeries (i.e. PA) scenarios on the system performance. Additionally, the case of using the stylus or the finger as writing tool is considered.

For the study of the writing input impact in the system performance, the same three scenarios considered in the SVC-onGoing competition \cite{tolosana2021icdar} have been used:

\begin{itemize}
\item \textbf{Task 1:} analysis of office scenarios using the stylus as input.
\item \textbf{Task 2:} analysis of mobile scenarios using the finger as input.
\item \textbf{Task 3:} analysis of both office and mobile scenarios simultaneously.
\end{itemize}

For the development of the system, the training dataset of the DeepSignDB database (1084 subjects) has been used. This means that the system has been trained using only signatures captured with a stylus writing tool. This will have a considerable impact on the system performance, as will be seen in Sect. \ref{sec:experimentalResults}. It is also important to highlight that, in order to consider a very challenging impostor scenario, the skilled forgery comparisons included in the evaluation datasets (not in the training ones) of both databases have been optimised using machine learning methods, selecting only the best high-quality forgeries.

\begin{figure}[t]
\centering
\includegraphics[width=1\linewidth]{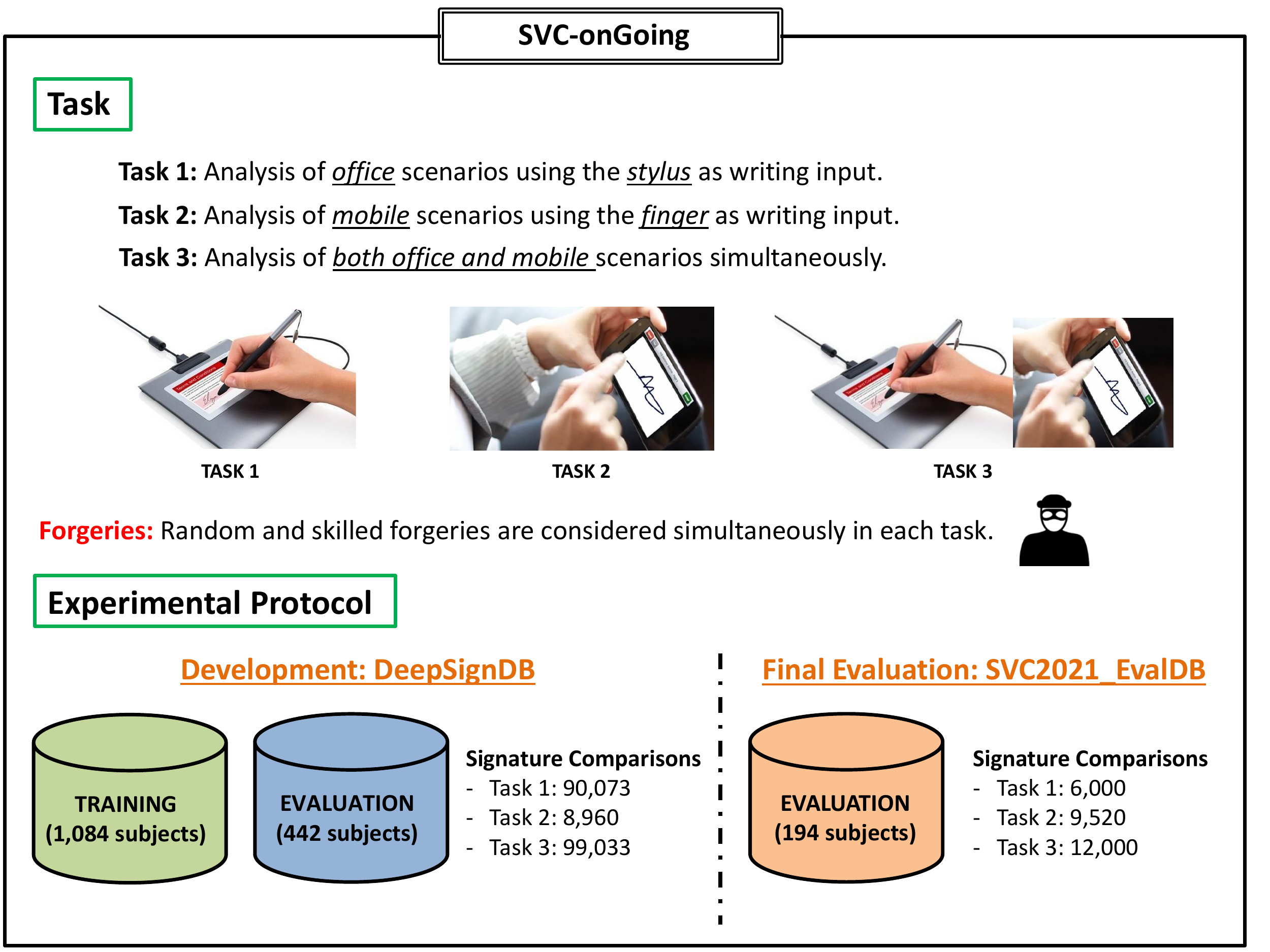}
\caption{Description of the tasks and experimental protocol details considered in SVC-onGoing Competition.} \label{SVCabstract}
\end{figure}

In addition, SVC-onGoing simulates realistic operational conditions \textbf{considering random and skilled forgeries simultaneously in each task.} A brief summary of the proposed experimental protocol used can be seen in Fig. \ref{SVCabstract}. For more details, we refer the reader to \cite{tolosana2021icdar}.

\subsection{Experimental Results}\label{sec:experimentalResults}

This section analyzes the results achieved in both DeepSignDB and SVC2021\_EvalDB databases.

\subsubsection{DeepSignDB}

\begin{figure}[t]
\centering
\subfigure[Task 1: Office Scenario]{
\includegraphics[width=0.49\linewidth]{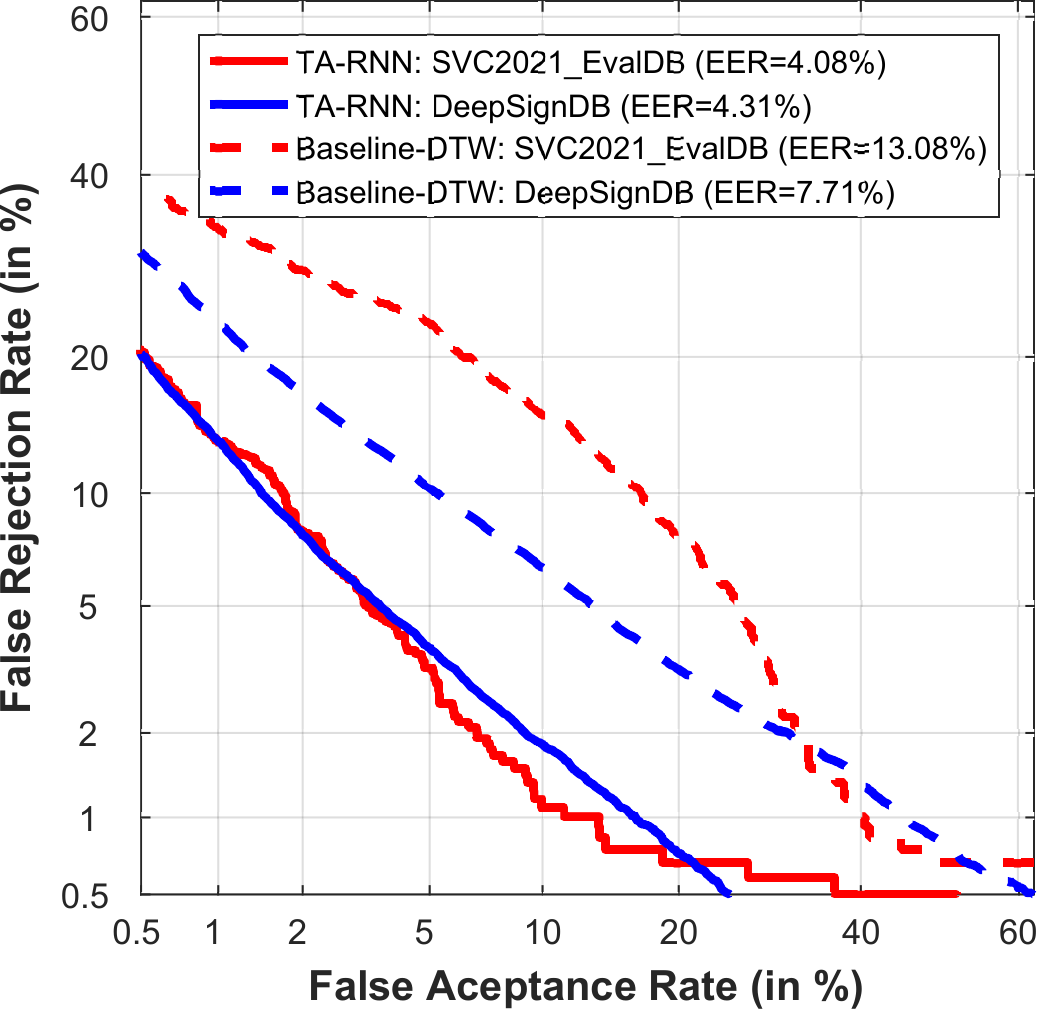}}
\subfigure[Task 2: Mobile Scenario]{
\includegraphics[width=0.49\linewidth]{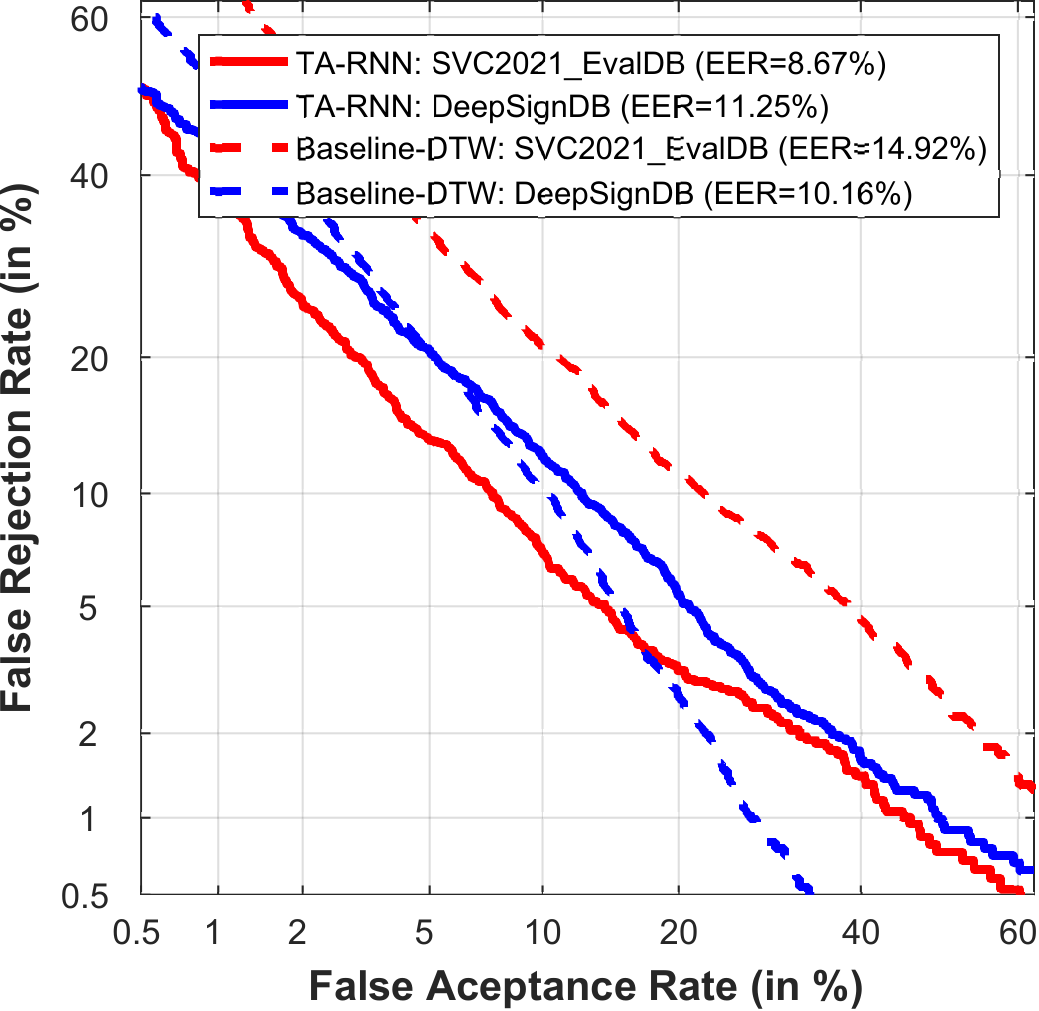}}
\subfigure[Task 3: Office/Mobile Scenario]{
\includegraphics[width=0.49\linewidth]{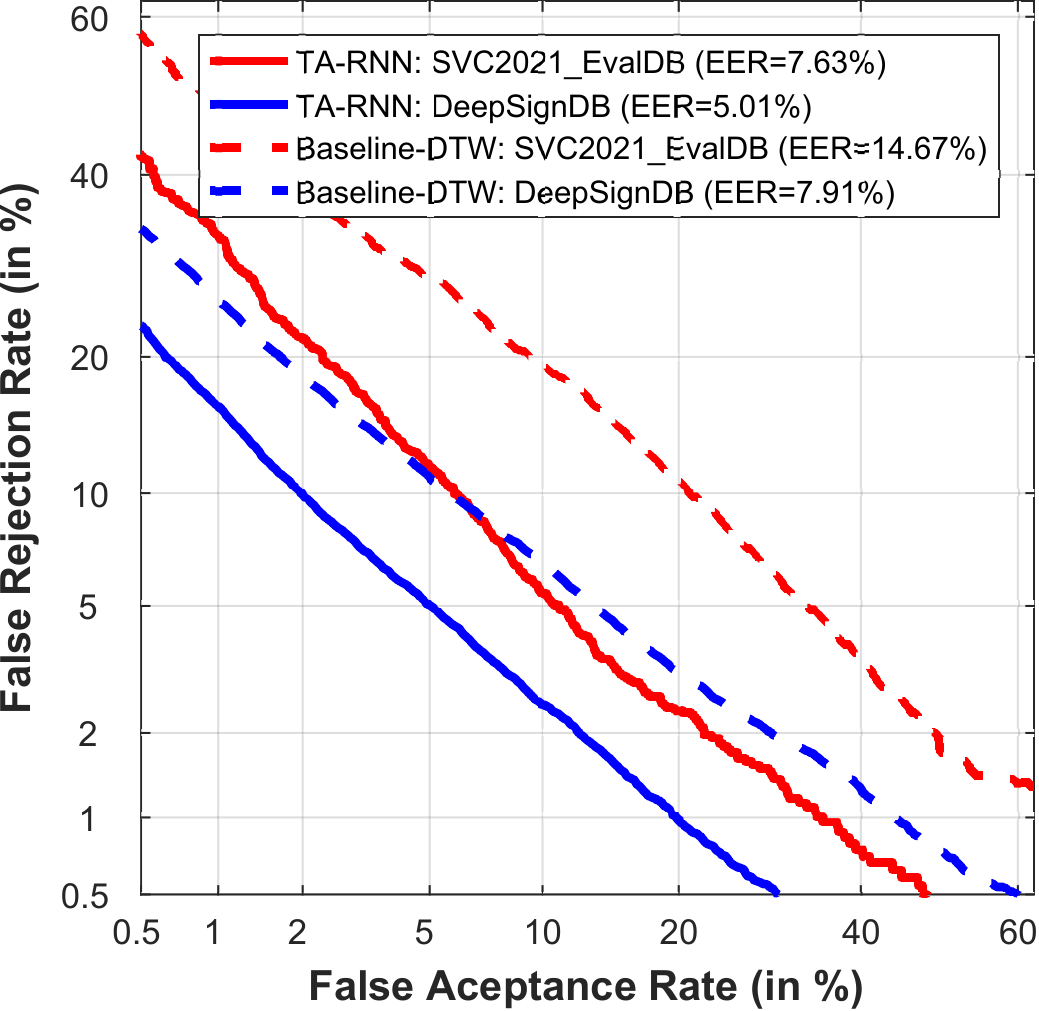}}
\caption{\textbf{Task Analysis:} Results in terms of DET curves over the evaluation dataset of DeepSignDB and SVC2021\_EvalDB for the three tasks considered.}
\label{DET_scenarios_1}
\end{figure}

In this first case, the evaluation dataset (442 subjects) of DeepSignDB was used to evaluate the performance of both DTW and TA-RNN systems. Fig. \ref{DET_scenarios_1} shows the results achieved in each of the three tasks using Detection Error Tradeoff (DET) curves and considering both random and skilled forgeries simultaneously. A Baseline DTW system (similar to the one described in \cite{Marcos_matching} based on X, Y spatial time signals, and their first- and second-order derivatives) is included in the image for a better comparison of the results. First, in all tasks we can see that the TA-RNN system has outperformed the traditional Baseline DTW. For Task 1, focused on the analysis of office scenarios using the stylus as writing input, the TA-RNN approach obtained a 4.31\% EER. Regarding Task 2, focused on mobile scenarios using the finger as writing input, a considerable system performance degradation is observed compared to the results of Task 1. In this case, the EER obtained was 11.25\%. This result proves the bad generalisation of the stylus model (Task 1) to the finger scenario (Task 2) as the model considered was trained using only signatures acquired through the stylus, not the finger. Finally, good results are generally achieved in Task 3 taking into account that both office and mobile scenarios are considered together, using both stylus and finger as writing inputs. The system obtained an EER of 5.01\%.

\subsubsection{SVC2021\_EvalDB}

In this section we present the results obtained in the evaluation of the novel \textit{SVC2021\_EvalDB} database. Similar to the previous section, we include in Fig. \ref{DET_scenarios_1} the Baseline DTW system.

It is important to highlight that TA-RNN achieves good EER results in the three tasks (4.08\%, 8.67\% and 7.63\% respectively) even if it is only trained with signatures introduced using the stylus as writing input. Also, it is interesting to compare the results achieved in each task with the results obtained using traditional approaches in the field (Baseline DTW). Concretely, for each of the tasks, the TA-RNN architecture achieves relative improvements of 68.81\%, 41.89\%, and 47.99\% EER compared to the Baseline DTW. These results prove the high potential of deep learning approaches such as TA-RNN for the on-line signature verification field, as commented in previous studies \cite{2021_TBIOM_DeepSign_Tolosana, tolosana2021deepwritesyn, lai2021synsig2vec2}.

Another key aspect to analyse is the generalisation ability of the proposed system against new users and acquisition conditions (e.g., new devices). This analysis is possible as different databases are considered in the development and final evaluation of the competition. Fig. \ref{DET_scenarios_1} show the results achieved using the DeepSignDB and SVC2021\_EvalDB databases, respectively. For Task 1, we can observe the good generalisation ability of the TA-RNN system, achieving results of 4.31\% EER for the development, and 4.08\% EER for the evaluation. Regarding Task 2, it is interesting to highlight that the TA-RNN system also obtains reasonable generalisation results. Similar trends are observed in Task 3.

Finally, for completeness, we also analyse the False Acceptance Rate (FAR) and False Rejection Rate (FRR) results of the proposed systems. Looking at Fig. \ref{DET_scenarios_1}, in general, for low values of FAR (i.e., high security), the TA-RNN system achieves good results in all tasks. It is interesting to remark that depending on the specific task, the FRR values for low values of FAR are very different. For example, analysing  a FAR value of 0.5\%, the FRR value is around 20\% for Task 1. However, the FRR value increases over 40\% for Task 2, showing the challenging conditions considered in real mobile scenarios using the finger as writing input. A similar trend is observed for low values of FRR (i.e., high convenience).

\subsubsection{Forgery Analysis}

\begin{figure}[!]
\centering
\subfigure[Task 1: Random Forgeries]{
\includegraphics[width=0.49\linewidth]{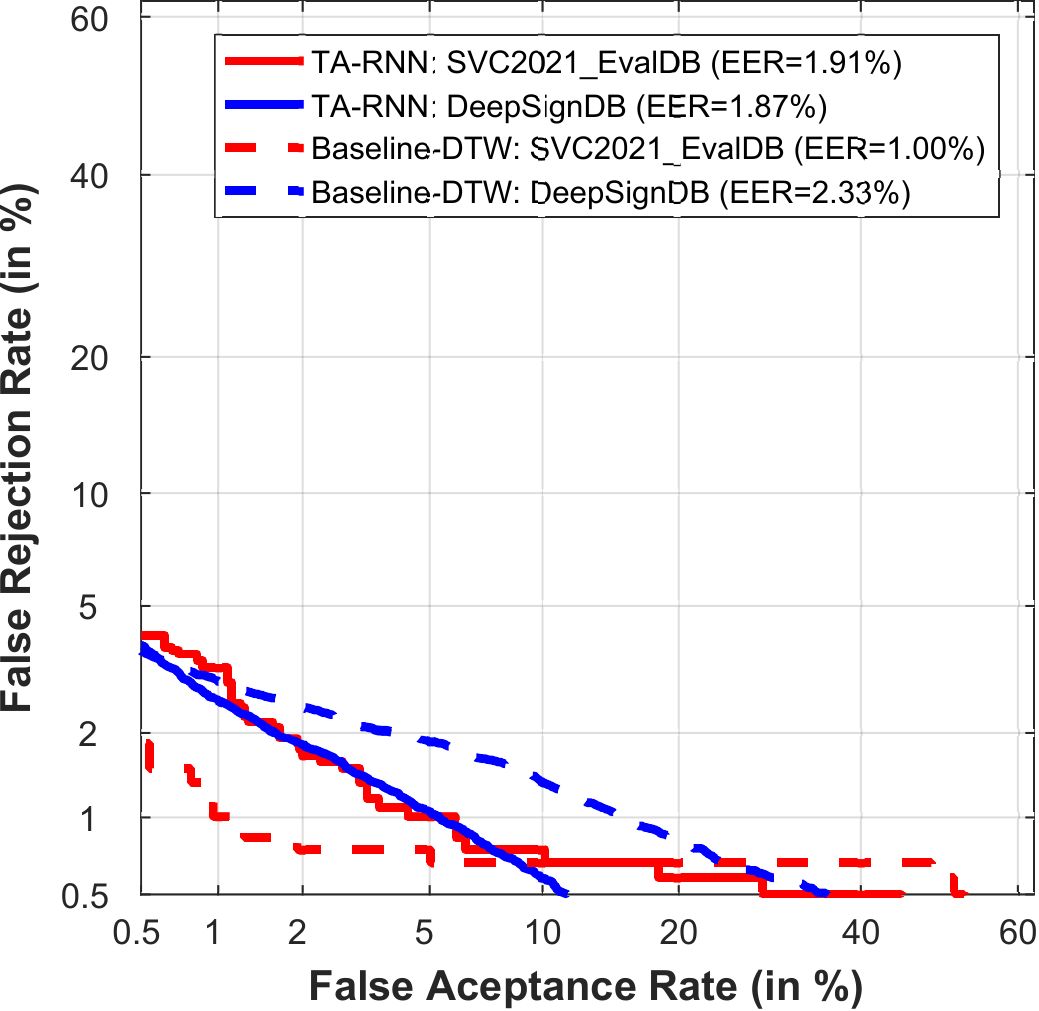}}
\subfigure[Task 1: Skilled Forgeries]{
\includegraphics[width=0.49\linewidth]{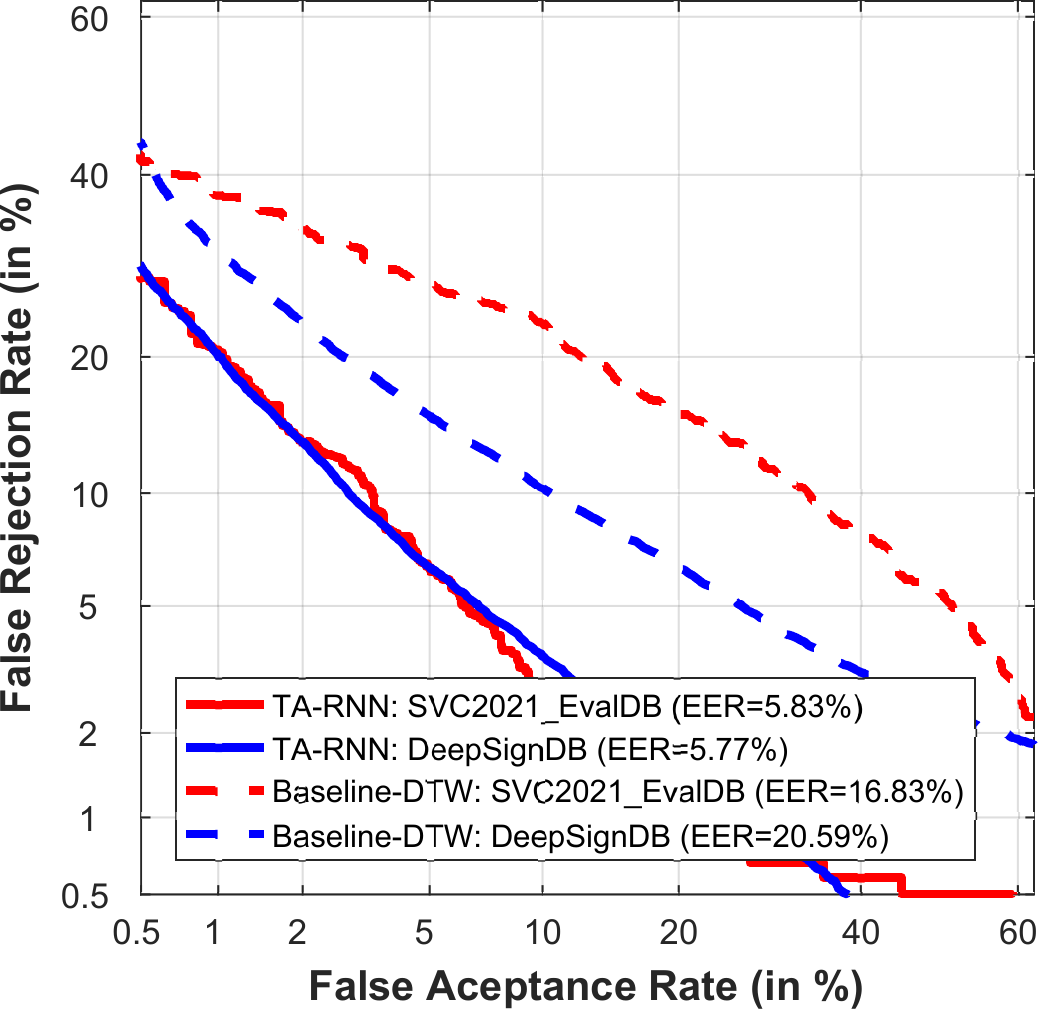}}
\subfigure[Task 2: Random Forgeries]{
\includegraphics[width=0.49\linewidth]{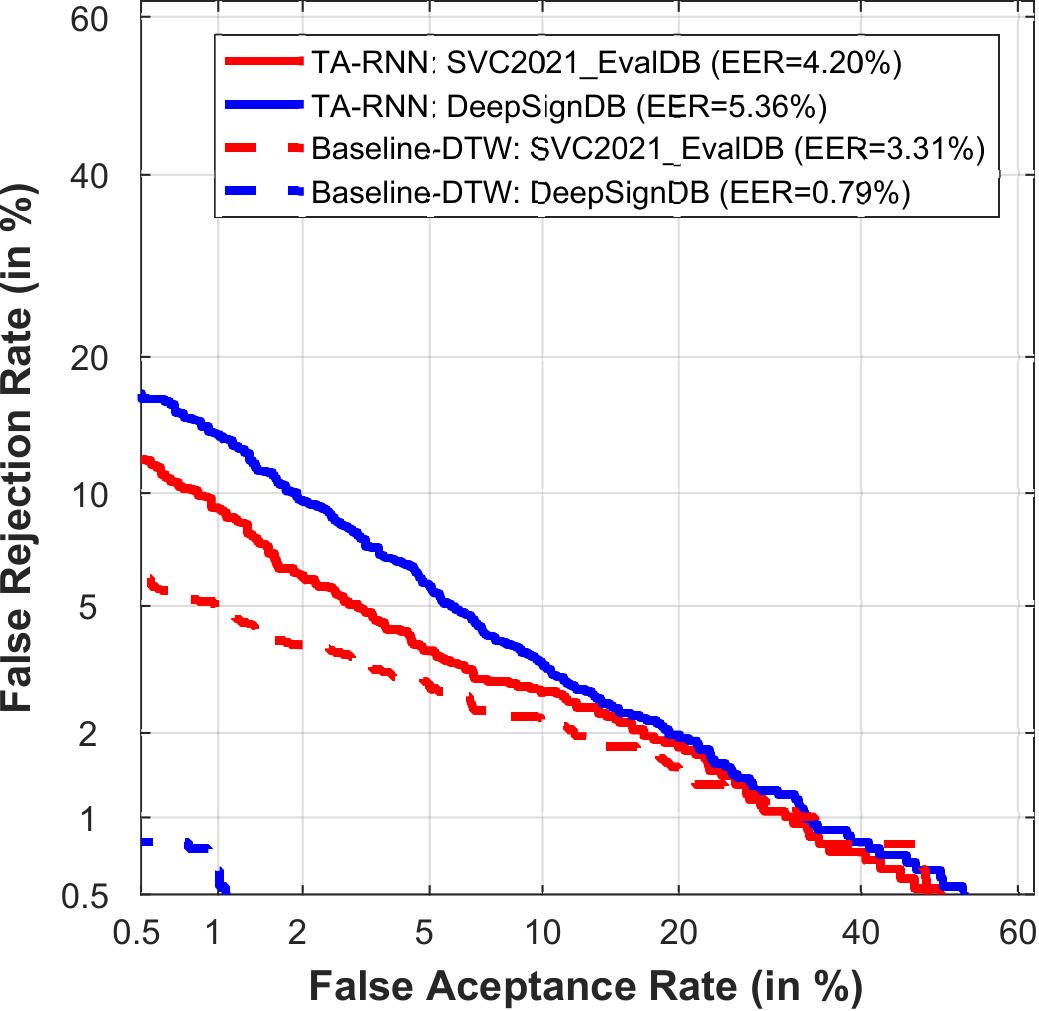}}
\subfigure[Task 2: Skilled Forgeries]{
\includegraphics[width=0.49\linewidth]{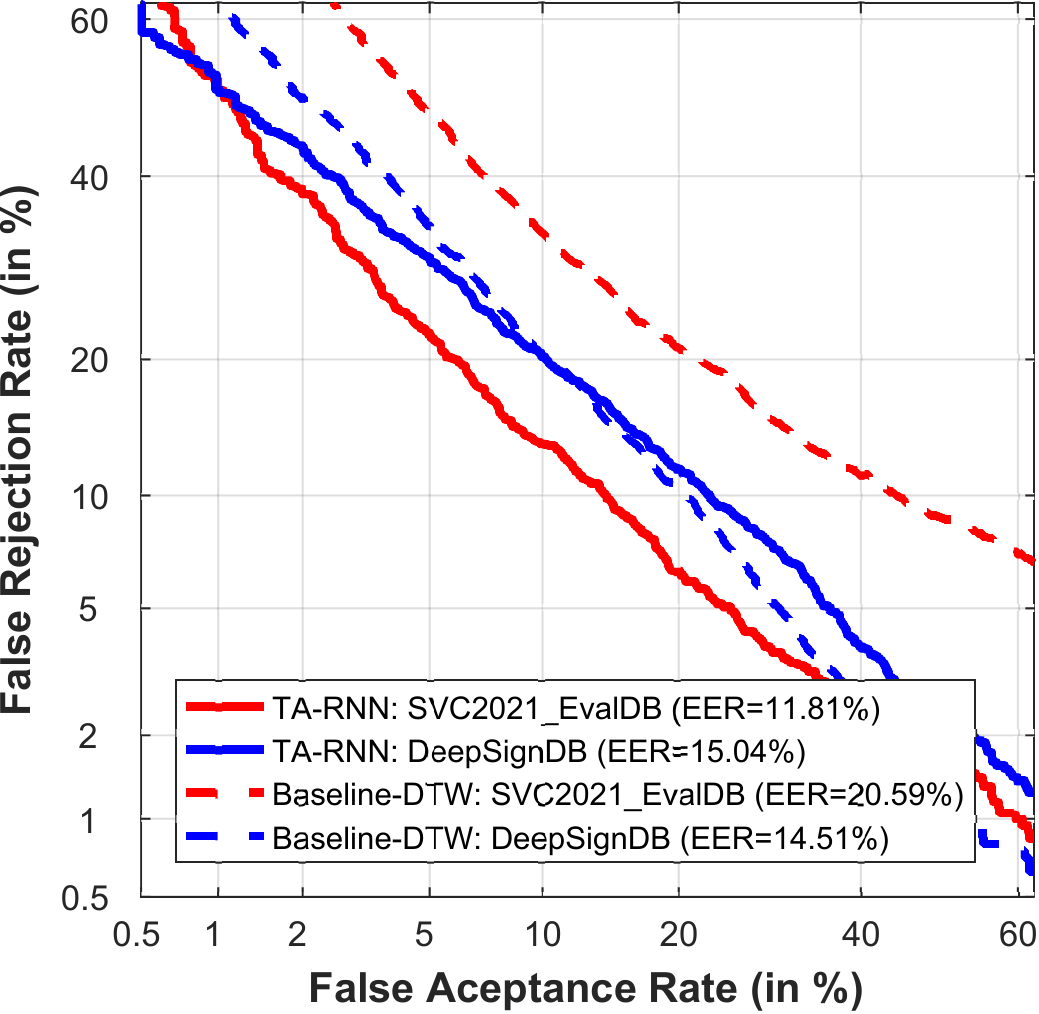}}
\subfigure[Task 3: Random Forgeries]{
\includegraphics[width=0.49\linewidth]{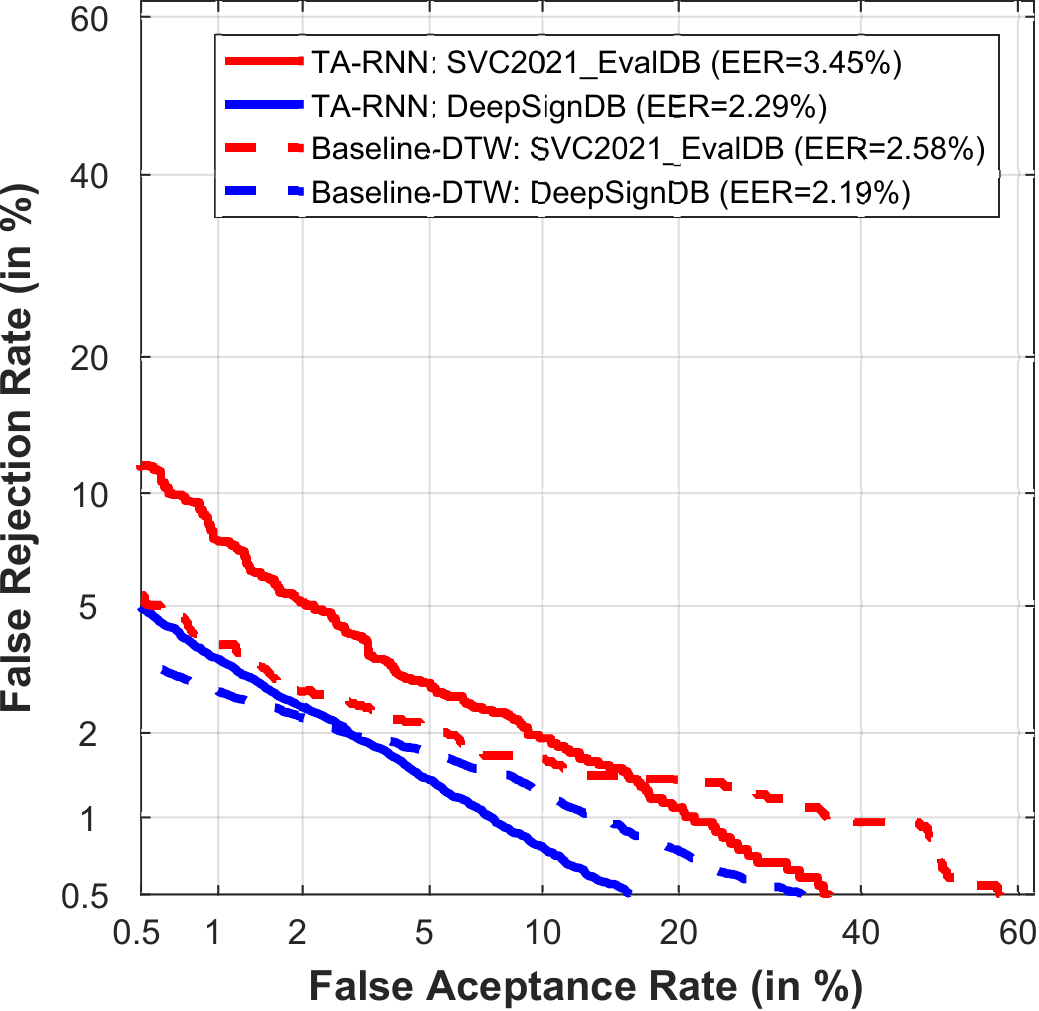}}
\subfigure[Task 3: Skilled Forgeries]{
\includegraphics[width=0.49\linewidth]{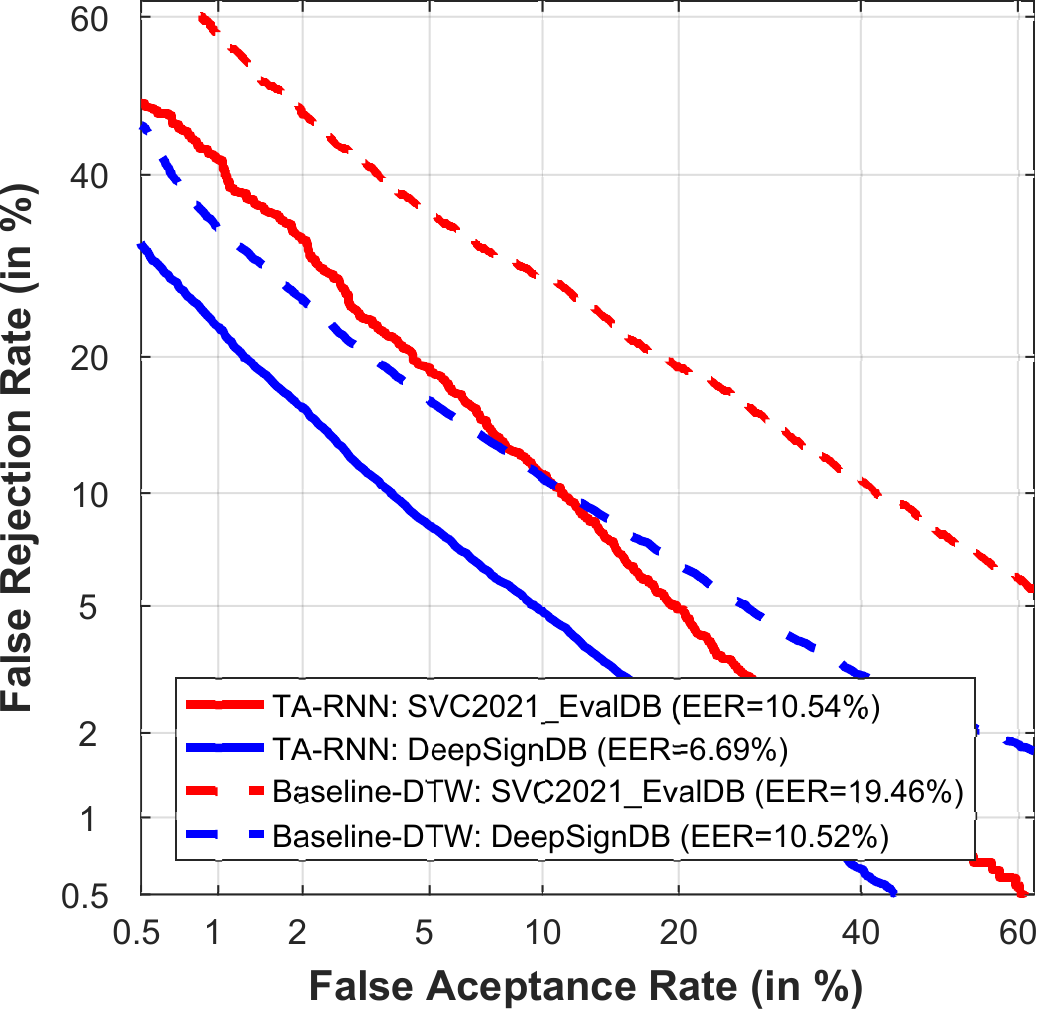}}
\caption{\textbf{Forgery Analysis:} Results in terms of DET curves over the evaluation dataset of DeepSignDB and SVC2021\_EvalDB for the three tasks and both types of forgeries separately.}
\label{DET_scenarios_2}
\end{figure}

This section analyzes the impact of the type of forgery in the proposed on-line signature verification system. In the evaluation of SVC-onGoing, both random and skilled forgeries are considered simultaneously in order to simulate real scenarios. Therefore, the winner of the competition was the system that achieved the highest robustness against both types of impostors at the same time \cite{tolosana2021icdar}. We now analyse the level of security of the two systems considered for each type of forgery, i.e., random and skilled. Fig. \ref{DET_scenarios_2} shows the DET curves of each task and type of forgery, including also the EER results, over both DeepSignDB and SVC2021\_EvalDB databases.

Analysing the skilled forgery scenario (Fig. 6b, 6d, and 6f), in all cases the TA-RNN system achieves the best results in terms of EER, outperforming the traditional Baseline DTW system in both SVC2021\_EvalDB and DeepSignDB databases.

Regarding the random forgery scenario, interesting results are observed in Fig. 6a, 6c, and 6e. In general, the TA-RNN system obtains worse results in terms of EER compared to the Baseline DTW system, which obtains EER results of 1.00\%, 3.31\%, and 2.58\% (SVC2021\_EvalDB) and 2.33\%, 0.79\% and 2.19\% (DeepSignDB) for each of the corresponding tasks of the competition, proving the potential of DTW for the detection of random forgeries. A similar trend was already discovered in previous studies in the literature \cite{2018_IEEEAccess_RNN_Tolosana}, highlighting also the difficulties of deep learning models to detect both skilled and random forgeries simultaneously.

Finally, seeing the results included in Fig. \ref{DET_scenarios_2}{}, we also want to highlight the very challenging conditions considered in SVC-onGoing compared with previous international competitions. This is produced mainly due to the real scenarios studied in the competition, e.g., several acquisition devices and types of impostors, large number of subjects, etc.

\section{Conclusions}\label{sec:conclusions}
This chapter carries out an analysis of Presentation Attack (PA) scenarios for on-line handwritten signature verification. Unlike traditional PAs, which use physical artefacts (e.g. gummy fingers and fake masks), the most typical PAs in signature verification represent an impostor interacting with the sensor in a very similar way followed in a normal access attempt (i.e., the PA is a handwritten signature, in this case trying to imitate to some extent the attacked identity). In a typical signature verification PA scenario, the level of knowledge that the impostor has and uses about the signature being attacked, as well as the effort and the ability to perform the forgeries, results crucial for the success rate of the system attack. 

The main contributions of this chapter are: 1) a brief overview of representative methods for PAD in signature biometrics; 2) the description of the different levels of PAs existing in on-line signature verification regarding the amount of information available to the impostor, as well as the training, effort and ability to perform the forgeries; and 3) analysis of system performance evaluation in signature biometrics under different PAs and writing tools considering new and publicly available signature databases.

Results obtained for both DeepSignDB and SVC2021\_EvalDB publicly available databases show the high impact on the system performance regarding not only the level of information that the attacker has but also the training and effort performing the signature. For the case of users using the finger as the writing tool, a recommendation for the usage of signature verification on smartphones on mobile scenarios (i.e., sitting, standing, walking, indoors, outdoors, etc.) would be to protect themselves from other people that could be watching while performing their genuine signature, as this is more feasible to do in a mobile scenario compared to an office scenario. This way skilled impostors (i.e. PA impostors) might have access to the global image of the signature but not to the dynamic information and system performance would be much better. This work is in line with recent efforts in the Common Criteria standardization community towards security evaluation of biometric systems, where attacks are rated depending on, among other factors: time spent, effort, and expertise of the attacker; as well as the information available and used from the target being attacked \cite{entregable_BEAT}.

\section{Acknowledgments}
The chapter update for the 3rd Edition of the book has received funding from the European Union’s Horizon 2020 research and innovation programme under the Marie Sklodowska-Curie grant agreement No 860315 (PRIMA) and No 860813 (TRESPASS-ETN). Partial funding also from INTER-ACTION (PID2021-126521OB-I00 MICINN/FEDER), Orange Labs, and Cecabank.

\bibliographystyle{spmpsci}
\bibliography{references}

\ifthenelse{\equal{false}{\buildbook}}{
\printindex
\printglossary
\bibliographystyle{spmpsci}
\bibliography{references}
}

\end{document}